\documentclass{article}

\PassOptionsToPackage{numbers, compress}{natbib}

\usepackage[final]{neurips_2021}

\usepackage[utf8]{inputenc} 
\usepackage[T1]{fontenc}    
\usepackage{hyperref}       
\usepackage{url}            
\usepackage{booktabs}       
\usepackage{amsfonts}       
\usepackage{xfrac}          
\usepackage{nicefrac}       
\usepackage{microtype}      
\usepackage{xcolor}         
\usepackage{acronym}  
\usepackage{amsmath}  
\usepackage{algorithm}
\usepackage{algorithmic}
\usepackage{graphicx}
\usepackage{subcaption}

\let\OLDthebibliography\thebibliography
\renewcommand\thebibliography[1]{
  \OLDthebibliography{#1}
  \setlength{\parskip}{0pt}
  \setlength{\itemsep}{1pt plus 0.3ex}
}

\title{Reinforcement Learning with Feedback from Multiple Humans with Diverse Skills}

\author{%
Taku Yamagata\\
Department of Engineering Mathematics\\
University of Bristol\\
Bristol, BS8 1TW \\
\texttt{taku.yamagata@bristol.ac.uk} \\
\And
Ryan McConville\\
Department of Engineering Mathematics\\
University of Bristol\\
Bristol, BS8 1TW \\
\texttt{ryan.mcconville@bristol.ac.uk} \\
\And
Ra{\'u}l Santos-Rodr{\'i}guez\\
Department of Engineering Mathematics\\
University of Bristol\\
Bristol, BS8 1TW \\
\texttt{enrsr@bristol.ac.uk} \\
}

\setcitestyle{square}

\begin{document}

\maketitle

\begin{abstract}
A promising approach to improve the robustness and exploration in Reinforcement Learning is collecting human feedback and that way incorporating prior knowledge of the target environment. It is, however, often too expensive to obtain enough feedback of good quality. To mitigate the issue, we aim to rely on a group of multiple experts (and non-experts) with different skill levels to generate enough feedback. Such feedback can therefore be inconsistent and infrequent. In this paper, we build upon prior work -- \textit{Advise}, a Bayesian approach attempting to maximise the information gained from human feedback -- extending the algorithm to accept feedback from this larger group of humans, the \textit{trainers}, while also estimating each trainer's reliability. We show how aggregating feedback from multiple trainers improves the total feedback's accuracy and make the collection process easier in two ways.
Firstly, this approach addresses the case of some of the trainers being adversarial. Secondly, having access to the information about each trainer reliability provides a second layer of robustness and offers valuable information for people managing the whole system to improve the overall trust in the system. It offers an actionable tool for improving the feedback collection process or modifying the reward function design if needed. 
We empirically show that our approach can accurately learn the reliability of each trainer correctly and use it to maximise the information gained from the multiple trainers' feedback, even if some of the sources are adversarial.
\end{abstract}

\section{Introduction}
\label{sec:intro}
The main goal of \ac{RL} is to design agents that learn a desired sequence of actions autonomously from the interactions with an environment. It is, however, often challenging to learn purely from these interactions, especially in the case that the environment has a high dimensionality and/or the length of the action sequence is long. In such cases, learning usually requires too many samples (interactions) before converging, preventing applying \ac{RL} to many applications. One promising approach to mitigate the issue is to incorporate prior knowledge of the target environment via human feedback. However, accurate feedback can be very costly to obtain.

In this work, we consider scenarios where it is possible to get feedback from a pool of multiple diverse experts / trainers (and potentially non-experts), whose feedback could be infrequent and inaccurate. Our proposed method estimates each trainer’s reliability, as it receives the feedback, and exploits them in the feedback aggregation process. The aggregated feedback is subsequently used to shape the \ac{RL} policy to improve the agent's behaviour.

Importantly, estimating each trainer's reliability also provides very useful knowledge for the people managing and overlooking these systems, increasing the overall robustness by offering better actionable insights into their inner workings. For example, if the reliability is found to be low for some of the trainers, this can be traced down to issues with either the trainers or the \ac{RL} setup. If the former, the manager could then inform (or replace) the trainers in order to improve the quality of their feedback or, if the latter, the manager could review the \ac{RL} setup -- reward function, state and action space etc. -- to make the behaviour based on the \ac{RL} setup match the trainers' feedback. We argue that having access to this information, with both managers and trainers in the loop, provide an extra safety layer in the deployment in practice of RL agents.

\paragraph{Related work} The idea of using human feedback in the \ac{RL} framework to accelerate its learning speed has been explored before. There are mainly two ways of influencing the agent's learning. One modifies (`shapes') reward signals, while the other modifies the agent's policy directly.
Most methods to incorporate human feedback modify a reward signal from the environment~\cite{Knox2012,Pilarski2011,Tenorio2010}. For example, Knox et al. proposed the TAMER framework, which assumes that an agent has a table for each state-action pair $\hat{H}(s,a)$ to accumulate the human feedback\cite{Knox2012}, and use it to generate the pseudo-rewards. 
However, it is in general difficult to come up with a reward value based on human feedback, such as ``yes, it is correct'' or ``no, it is wrong''. 
Also, recent papers have observed that a more effective use of human feedback is to directly inform on concrete policies~\cite{Knox2012, Thomaz2008}.
Within this remit, Griffith et al. proposed \textit{Advise}\cite{Griffith2013}, which estimates the Bayes optimum feedback policy and directly applies it to an agent's policy.
The authors show that it outperforms existing approaches and it is robust against infrequent and inconsistent feedback. Outside RL, the problem of using feedback (labels) coming from multiple trainers (or annotators in this case), has been addressed under the learning from crowds paradigm for both non-sequential~\cite{raykar} and sequential data~\cite{ni}. In these works, the idea is both to obtain accurate aggregate feedback as well as estimating the reliability of the trainers. 

\paragraph{Structure} In this work, we build upon these ideas to propose a method that allows RL agents to learn from the received feedback from multiple trainers with unknown skill levels while estimating each of the trainer's reliability in an online fashion. The main contribution is presented in Section \ref{sec:method}. The empirical evaluation of the approach is shown in Section \ref{sec:experiments}. Finally, Section \ref{sec:conclusion} depicts the main conclusions and future research lines.

\section{Method}
\label{sec:method}
In order to incorporate feedback into the RL algorithm, we build upon the \textit{Advise} algorithm~\cite{Griffith2013} and thus provide a brief description of the approach. 
\textit{Advise} assumes binary feedback from a trainer that returns either `right' or `wrong' for a particular agent's choice of action. 
The feedback is accumulated for each state-action pair separately, and it is used to derive a trainer's policy, denoted by $\pi_{F}\left(s,a\right)$, which is then used to modify the agent's policy. 
Additionally, $C$ is defined as the probability that the trainer gives the right (consistent) feedback, and assuming a binomial distribution, the trainer's policy is as follows.
\begin{equation} \label{eq220}
\pi_{F}\left(s,a\right) = C^{\Delta(s,a)} (1-C)^{\sum_{j\neq a}\Delta(s,j)}
\end{equation}
where $\Delta(s,a)$ is the difference between the amount of positive and negative feedback from the trainer. 
The policy of the trainer is combined with $\pi_{R}\left(s,a\right)$ (policy from the underlying RL  algorithm) by multiplying them together so that the final policy becomes as
\begin{equation} \label{eq222}
\pi \left(s,a\right) \propto \pi_{F}\left(s,a\right) \cdot \pi_{R}\left(s,a\right).
\end{equation}
While \textit{Advise} provides a mechanism for incorporating feedback, it has two major limitations: it requires the consistency level ($C$) prior to receiving feedback, which may be unknown or difficult to estimate in many applications.  
The second limitation is the restriction to a single trainer. 

When multiple trainers are available, it may be beneficial to incorporate feedback from all sources.
Further, by estimating their reliability, reliable trainer feedback may be incorporated while avoiding adversarial effects from unreliable trainers.
Thus, in this work, we extend \textit{Advise} to take multiple trainers feedback into account while treating the consistency level as an unknown parameter, estimating it in an online fashion. 

\subsection{Consistency level estimation}
We propose the following method to estimate the $n_{th}$ trainer's consistency level ($C_{[n]}$). The estimation has two steps, and the first step produces an estimate of the consistency level for a given state and action pair ($C_{[n] s,a}$), and then takes an average to obtain a universal $C_{[n]}$ for all state-action pairs. 
In the current and following subsections, all discussions are regarding a single $n_{th}$ trainer, hence we omit the trainer's index $[n]$ from all the variables for simplicity. 

We consider estimating the trainer's consistency level for a given state action pair ($C_{s,a}$). With a given amount of positive feedback ($h_{s,a}^{+}$) and negative feedback ($h_{s,a}^{-}$) on a given state action pair, this can be derived by maximizing the following log-likelihood function which has a marginalised out hidden parameter $\mathcal{O}_{s,a}$,
\begin{equation} \label{eq302}
l(C_{s,a})=\log\left( \sum_{\mathcal{O}_{s,a}} p\left( h_{s,a}^{+}, h_{s,a}^{-}, \mathcal{O}_{s,a};C_{s,a}\right)\right).
\end{equation}
The hidden parameter is a boolean that is 1 when $a$ is the optimal action at state $s$, and 0 when it is not.
We then use the Expectation-Maximization (EM) algorithm~\cite{Dempster1977} to compute a maximum likelihood estimate of the consistency level ($C_{s,a}$). The $i_{th}$ iteration of the M-step can be written as follows,
\begin{equation} \label{eq304}
C_{s,a}^{\left(i+1\right)} = \frac{P_{1} \cdot h_{s,a}^{+} + P_{0} \cdot h_{s,a}^{-}}{h_{s,a}^{+} + h_{s,a}^{-}},
\end{equation}
where $C_{s,a}^{\left(i+1\right)}$ is the estimated consistency level at the $i_{th}$ iteration, $P_0$ and $P_1$ are given as follows:
\begin{equation} \label{eq305}
\begin{split}
P_0 = P\left( \mathcal{O}_{s,a}=0 | h_{s,a}^{+}, h_{s,a}^{-};C_{s,a}^{\left(i\right)}\right), \\
P_1 = P\left( \mathcal{O}_{s,a}=1 | h_{s,a}^{+}, h_{s,a}^{-};C_{s,a}^{\left(i\right)}\right).
\end{split}
\end{equation}
The E-step fundamentally requires computing Eq.~\ref{eq305}, using Eq.~\ref{eq220} and the probabilities derived from interaction with the environment. $P_{1}^{Q}(s,a)$ is the probability of the optimal action derived from the interaction. 
As a result, they can be written as follows.
\begin{equation} \label{eq307}
\begin{split}
P_0 &= \sum_{a'\neq a} P_{1}^{Q}(s,a') \cdot \left(C_{s,a'}^{(i)}\right)^{\Delta(s,a')}\left(1-C_{s,a'}^{(i)}\right)^{\sum_{j\neq a'}\Delta(s,j)} / Z, \\
P_1 &= P_{1}^{Q}(s,a) \cdot \left(C_{s,a}^{(i)}\right)^{\Delta(s,a)} \left(1-C_{s,a}^{(i)}\right)^{\sum_{j\neq a}\Delta(s,j)} / Z.
\end{split}
\end{equation}
We set $P_{1}^{Q}(s,a)=\pi_R(s,a)$ 
and $Z$ is the partition function which is sum of the numerators of the equations.
Once it computes the consistency level for each state-action pair, we average it over state-action and come up with one consistency level for each trainer. It is sensible to take an average over state-action pairs that are actually being experienced with the current policy. We use the following recursive averaging method over the experiencing state-action pairs.
\begin{equation} \label{eq310}
C' = C + \alpha \cdot ( C(s,a) - C ),
\end{equation}
where $\alpha \in [0,1]$ is the learning rate, $C$ and $C'$ are averaged consistency levels before and after the update respectively. $C(s,a)$ is the estimated consistency level for the current state-action pair.
\subsection{Adaptive learning rate}
We consider an approach which adaptively changes the learning rate $\alpha$ so that the updated consistency level $C'$ has maximum accuracy. We define the generalised variance $\sigma'^2$ and the precision $\lambda'$ of $C'$ as follows:
\begin{equation} \label{eq:prec_def1}
\sigma'^2 = 1/\lambda' = \mathbb{E}\left[\left(C' - C^*\right)^2\right],
\end{equation}
and similarly for $C$ and $C(s,a)$,
\begin{equation} \label{eq:prec_def2}
\begin{split}
\sigma^2 = 1/\lambda &= \mathbb{E}\left[\left(C - C^*\right)^2\right] \\
\sigma_{s,a}^2 = 1/\lambda_{s,a} &= \mathbb{E}\left[\left(C(s,a) - C^*\right)^2\right],
\end{split}
\end{equation}
where $C^*$ is the true consistency level. By substituting Eq.~\ref{eq310} to Eq.~\ref{eq:prec_def1} and computing $\frac{\partial\sigma'^2}{\partial\alpha}=0$, we obtain the value of $\alpha$ that minises the $\sigma'^2$ as follows:
\begin{equation} \label{eq:adap_lr_alpha}
\alpha = \frac{\lambda_{s,a}}{\lambda_{s,a}+\lambda}.
\end{equation}

With the learning rate in place, the precision of the updated consistency level $\lambda'$ becomes:
\begin{equation} \label{eq:adapt_lr_lambda}
\lambda' = \lambda_{s,a}+\zeta\lambda,
\end{equation}
where $\zeta$ is a decay factor (hyper-parameter) and it should be 1.0 in theory; however, it could become less than one when the true value of $C$ changes over time or the $\lambda_{s,a}$ estimate (described in below) is imperfect. $\zeta$ is introduced to prevent the learning rate from becoming too small, which will cause halting the update. We set $\zeta=0.98$ in our  experiments.
With Eq.~\ref{eq:adap_lr_alpha} and Eq.~\ref{eq:adapt_lr_lambda}, we can compute $\alpha$ and $\lambda$ recursively by receiving $\lambda_{s,a}$. The procedure is summarised in the Appendix~\ref{app:adaptive_learning_rate_summary}.

Now, we consider how to compute $\lambda_{s,a}$.
The consistency level $C(s,a)$ estimation is based on two sources, namely, information from the underlying RL algorithm ($P_{1}^{Q}$) and the trainer's feedback ($h_{s,a}^{+}$ and $h_{s,a}^{-}$). Thus, we estimate these precisions separately and combine them by multiplying them together to determine the consistency level's precision. We do not need to obtain the absolute value of $\lambda_{s,a}$, as any scaling factors are cancelled out when we compute $\alpha$ in Eq.~\ref{eq:adap_lr_alpha}, as long as they are constant with respect to all $(s,a)$ over all time steps.

We use  the absolute value of state-action value function (Q function), added up over all actions as a metric for the underlying RL accuracy.
\begin{equation} \label{312}
\hat{\lambda}_Q = \sum_{a \in A} |Q(s,a)|.
\end{equation}
We use the number of feedbacks for the given state as a metric for the trainer's feedback accuracy, i.e.,
\begin{equation} \label{313}
\hat{\lambda}_{FB} = \sum_{a \in A} h_{s,a}^{+} + h_{s,a}^{-}.
\end{equation}
 The approximation of $\lambda_{s,a}$ is derived as,
\begin{equation} \label{314}
\lambda_{s,a} = \hat{\lambda}_Q \cdot \hat{\lambda}_{FB}.
\end{equation}
Although this form of $\lambda_{s,a}$ is empirically shown to work well, it can be further improved -- especially for $\lambda_Q$. The above $\lambda_Q$ only works when the value function is initialised with zero (or a very small value). This is left as future work.

\subsection{Multiple trainers}
 In order to incorporate multiple trainers, we assume each trainer has a different consistency level, and the $n_{th}$ trainer's consistency level is denoted by $C_{[n]}$. The Bayes optimal method to combine probabilities from (conditionally) independent sources is multiplying them together~\cite{Bailer-Jones2011}. Hence, the policy for overall multiple trainers $\pi_{F}(s,a)$ can be derived as follows by  employing each trainer's policy given in Eq.\ref{eq220},
\begin{equation}
\label{eq221}
\pi_{F}(s,a) \propto \prod_{n=1}^{N} \left(C_{[n]}\right)^{\Delta_{[n]}(s,a)}\left(1-C_{[n]}\right)^{\sum_{j\neq a}\Delta_{[n]}(s,j)}
\end{equation}
where $N$ is the number of trainers, and $\Delta_{[n]}(s,a)$ is the difference between positive and negative feedback on the state $s$ and action $a$ from the $n_{th}$ trainer. 

\section{Experiments}
\label{sec:experiments}
We used a 5x5 grid world Pac-Man to evaluate our approach because it is used in other Policy Shaping papers~\cite{Griffith2013,Cederborg2015}.
The goal is to eat all the food pellets without being caught by ghosts. Once clear the game, +500 reward is given while -500 reward is given if Pac-Man is caught by ghosts. Also, each pellet awards +10 points, and each time step costs 1 point (rewarded -1 point.)
The state representation includes Pac-Man's position, the position and orientation of the ghost, and each food pellet's presence.
Similar to \cite{Griffith2013}, we also use Oracle to simulate human feedback because it allows us to sweep parameters of feedback likelihood ($L$) that specifies how often a trainer gives feedback and consistency level ($C$). The Oracle was created by using Q-learning\cite{Watkins1989} on the environment before the experiments. We also use Q-Learning as the underlying \ac{RL} algorithm. Further details for the experiment are in Appendix~\ref{app:exp_details}.

Figure~\ref{fig:ST_1} shows results for multiple trainers. Part (a) illustrates the averaged total rewards for an example with eight trainers with uniformly distributed consistency levels between 0.2 to 0.9 and the feedback likelihood $L=0.2$. We show both the consistency level estimation on, and without the estimation (fixed $C=0.8$). The results show a clear benefit over the simple Q-Learning and the fixed consistency level results. Part (b) shows the consistency level estimations for each trainer, with all of them converging towards their true value.
\begin{figure}
\begin{subfigure}{.5\textwidth}
  \centering
  \includegraphics[width = 1.1\linewidth]{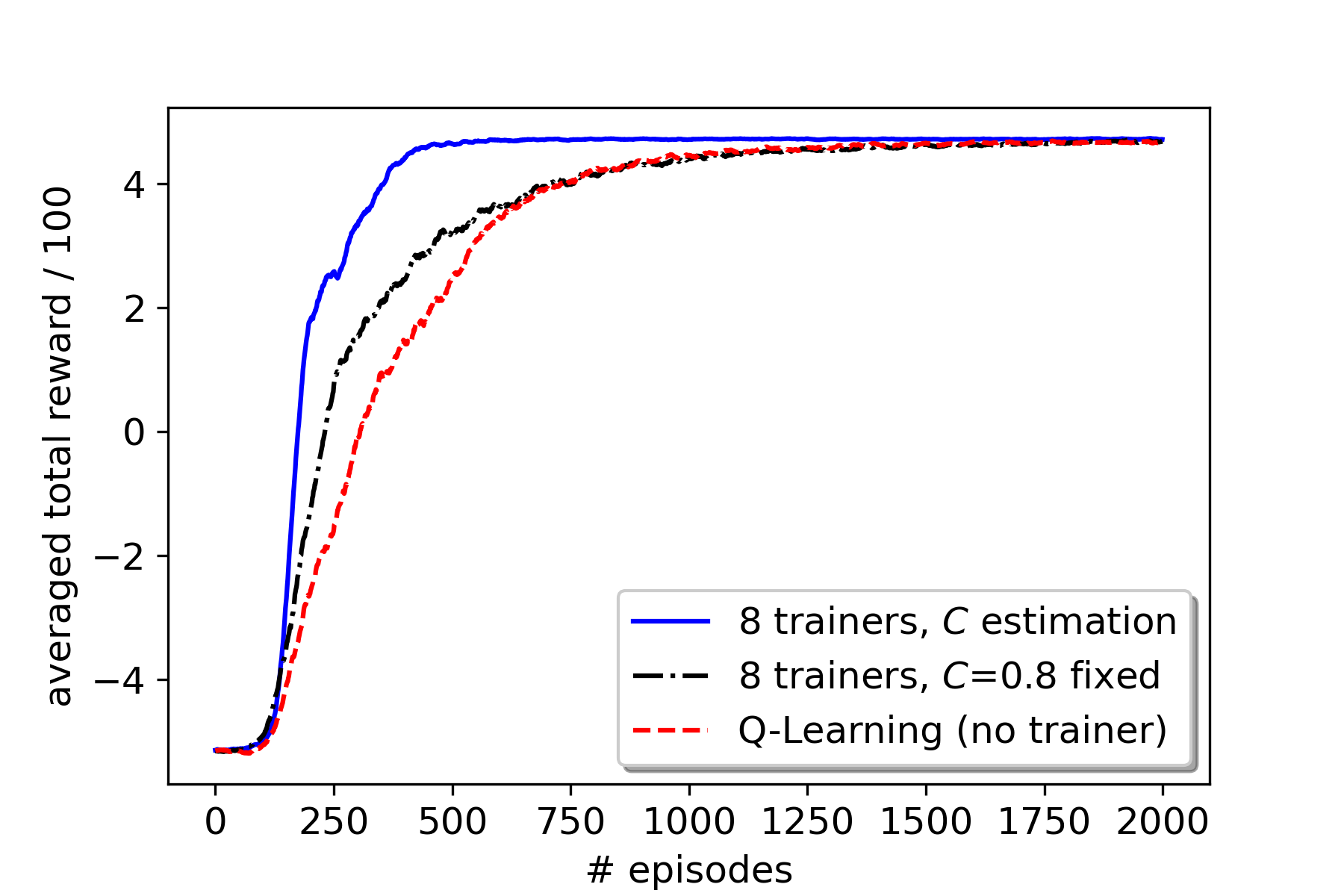}
  \caption{Averaged total rewards}
\end{subfigure}%
\begin{subfigure}{.5\textwidth}
  \centering
  \includegraphics[width = 1.1\linewidth]{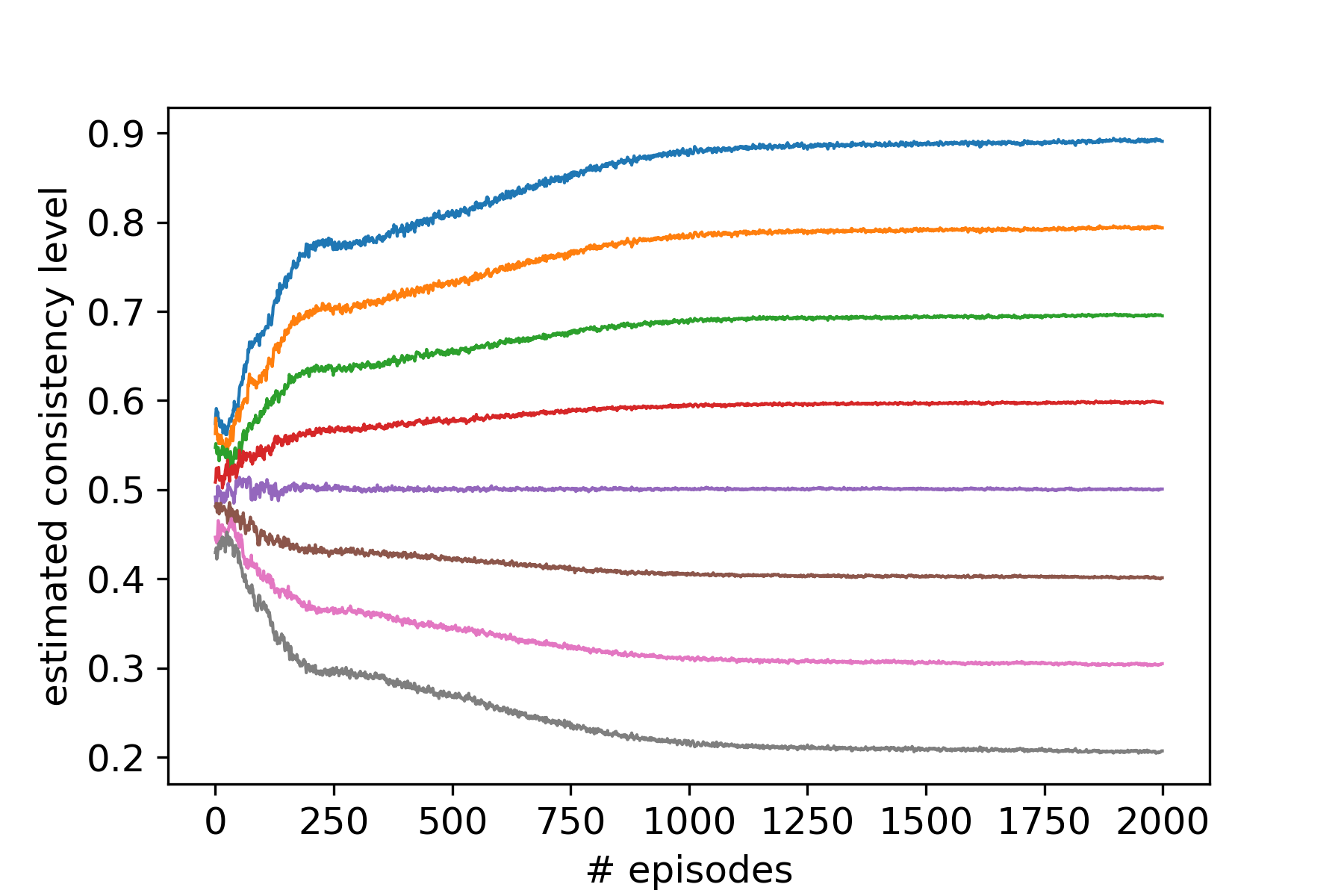}
  \caption{Estimated consistency levels for eight trainers}
\end{subfigure}
\caption{(a) Averaged total reward comparing fixed consistency ($C=0.8$) and estimating consistency level for eight trainers. The true consistency levels are uniformly distributed between 0.2 and 0.9. We also added a normal Q-Learning result for reference. (b) The estimated consistency levels for the eight trainers. It shows all converge to the right values.}\label{fig:ST_1}
\end{figure}

\begin{figure}
\begin{subfigure}{.5\textwidth}
  \centering
  \includegraphics[width = 1.1\linewidth]{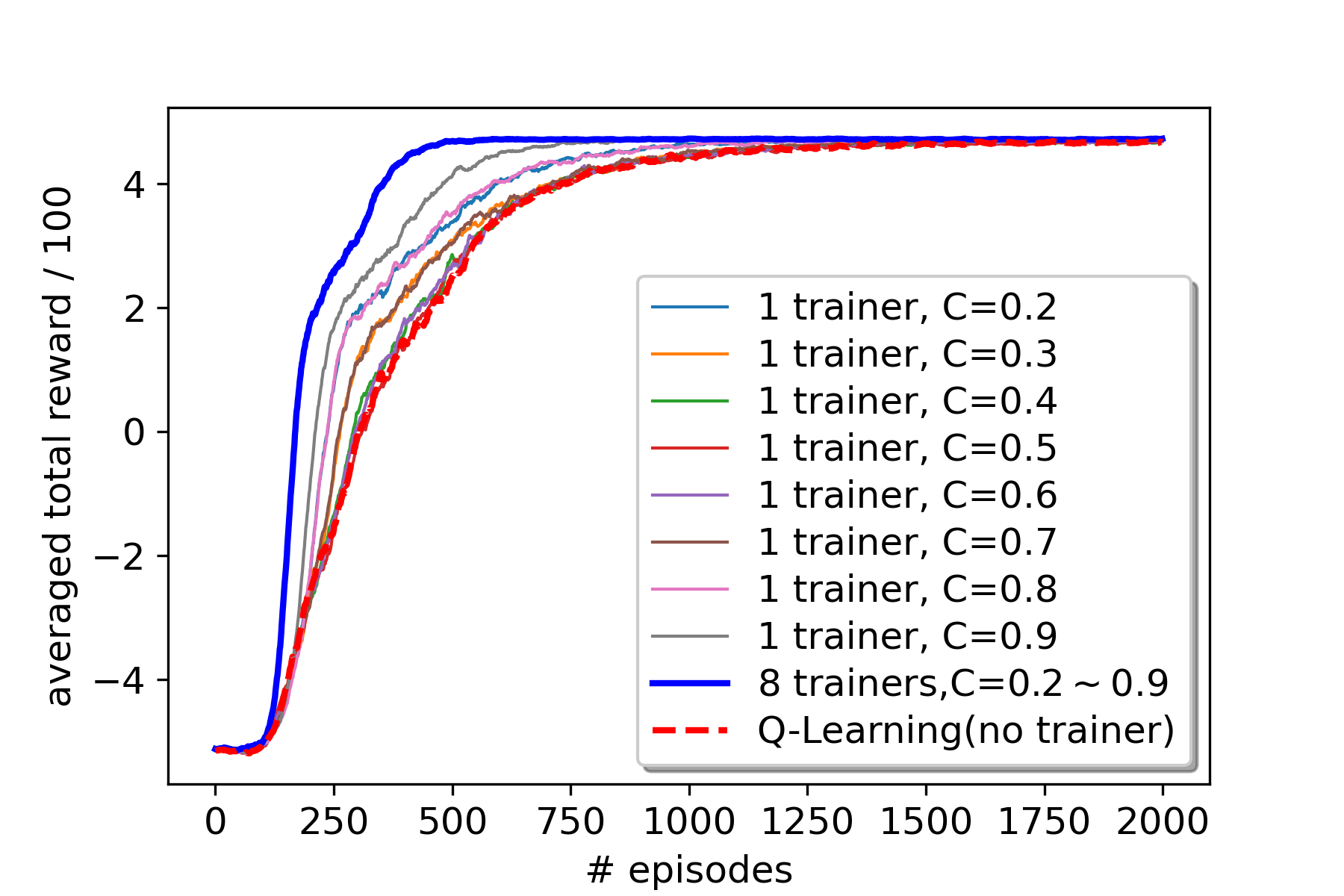}
  \caption{Single trainer case (overall)}
\end{subfigure}%
\begin{subfigure}{.5\textwidth}
  \centering
  \includegraphics[width=1.1\linewidth]{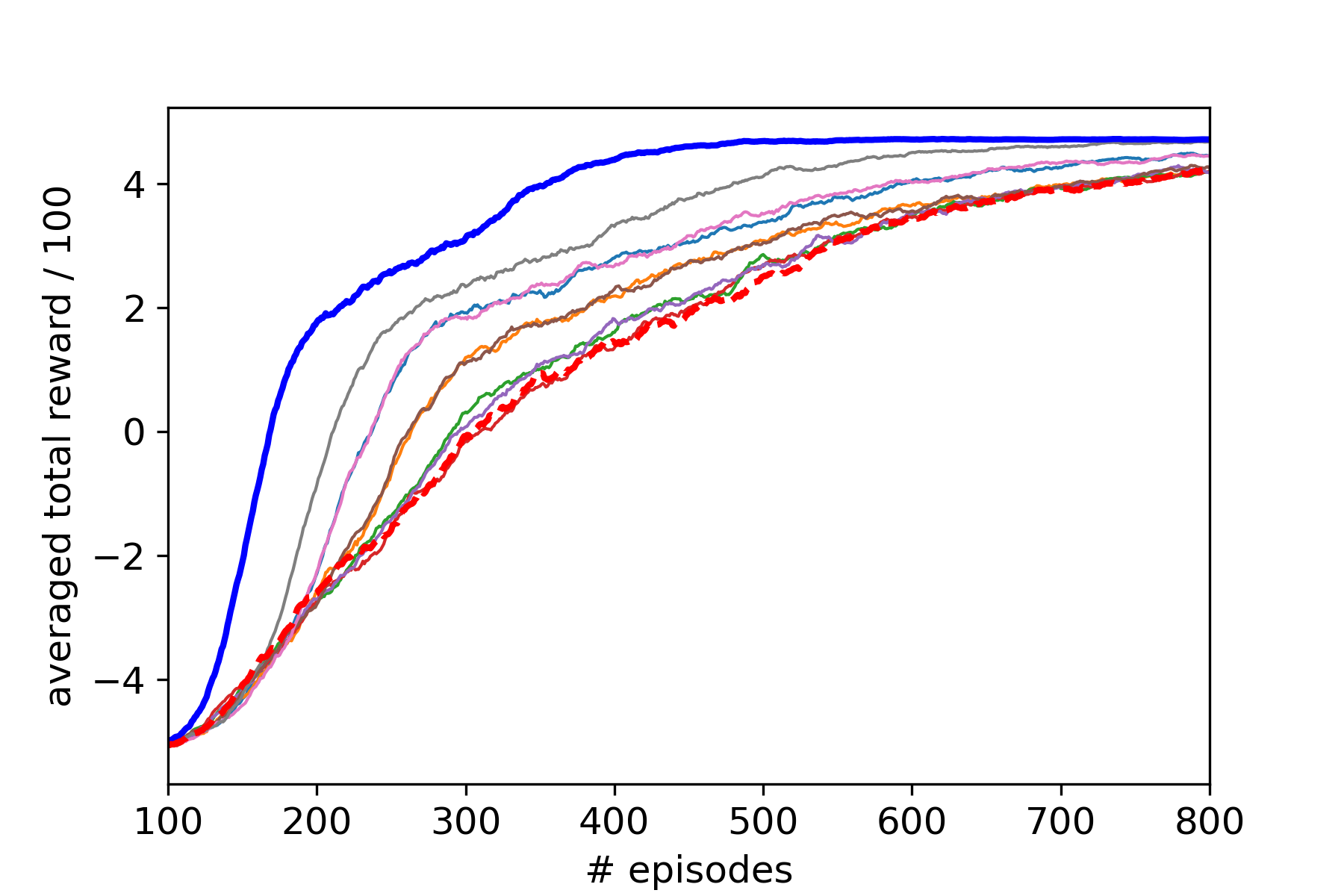}
  \caption{(focus on the early learning stage)}
\end{subfigure}
\caption{(a) Averaged total reward for the single trainer cases with various consistency levels. We also added a normal Q-Learning agent and the eight trainers' results for reference. (b) Same as Part(a), focusing on the early stages of the learning process (\#episodes=100 to 800) to show each result more clearly.}\label{fig:ST_2}
\end{figure}

\begin{figure}
  \centering
  \includegraphics[width = 0.76\linewidth]{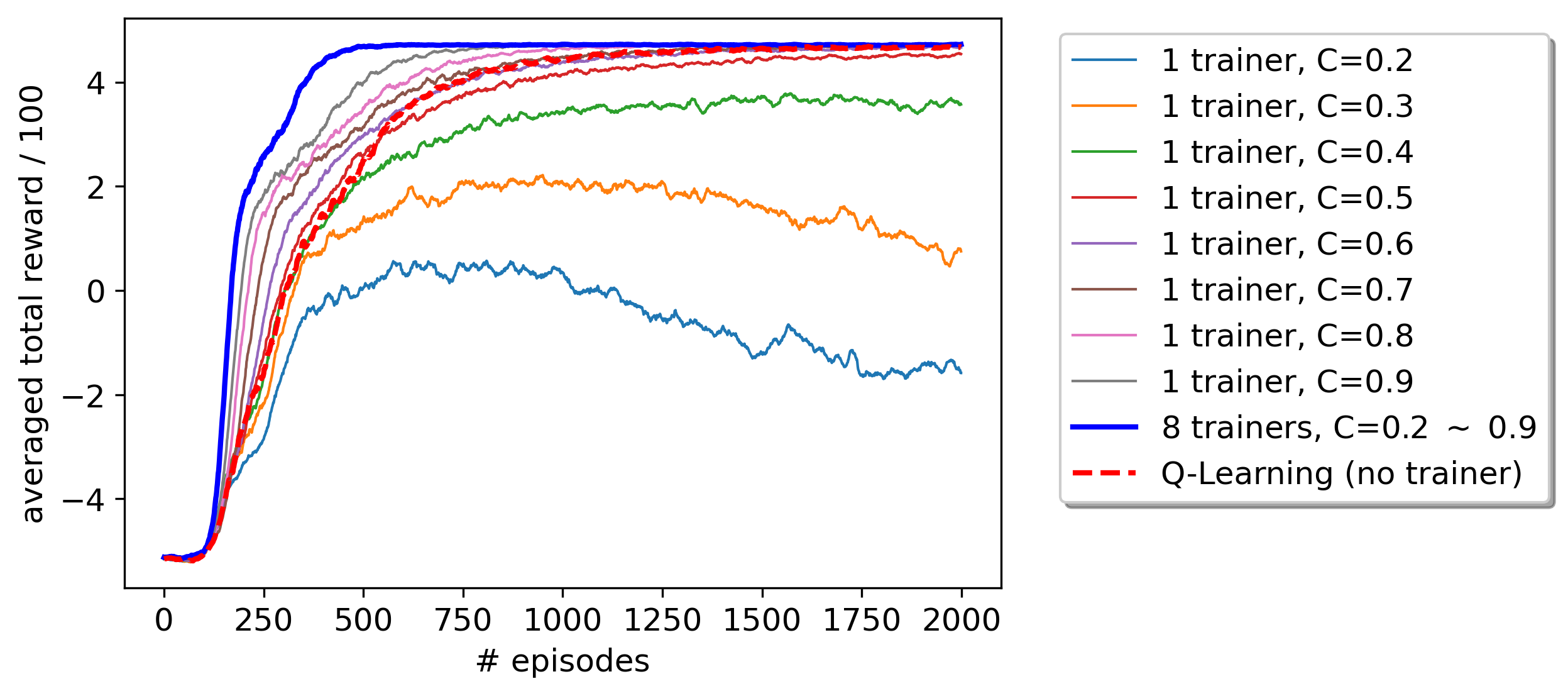}
  \caption{(a) Averaged total reward for the single trainer cases without estimating the consistency level (fixed to 0.8.) We also added a normal Q-Learning and the eight trainers with the consistency level estimation results for reference.}\label{fig:ST_3}
\end{figure}

We also simulate single trainer cases with various consistency levels and the feedback likelihood $L=0.2$ with the consistency level estimation. The result shows how much each trainer contributes to the overall learning process. 
Figure~\ref{fig:ST_2} Part (a) shows the total rewards for overall the X-axis range. Part (b) is an enlarged view of Part (a), focusing on the early part of the learning process. They indicate that the results with a trainer whose consistency level is farther from 0.5 are better. Also, the consistency levels that have the same distance from 0.5 -- such as 0.2 and 0.8 -- give very similar results. These are reasonable as the feedback is binary (`right' or `wrong'.) For example, a trainer who gives always incorrect feedback is as helpful as a trainer who always provide us with right feedback, once we know their consistency levels. 
We also evaluate the same single trainer cases without learning the consistency level (fixed $C=0.8$). The results (Fig.~\ref{fig:ST_3}) demonstrate that specifying an incorrect consistency level have a severe impact on the overall performance.

\section{Conclusion and Future Work}
\label{sec:conclusion}
We introduced a method to estimate the trainer's reliability (consistency) online and take multiple trainers' feedback to improve the robustness properties and exploration in \ac{RL}. It eases the feedback collection process and improves its accuracy even with adversarial trainers. Additionally the reliability estimates provide a second layer of robustness and helps with debugging any issues with the trainers and the \ac{RL} setups. Our experiments showed that it accelerated the learning speed over conventional Q-Learning, and achieved a learning speed close to the ideal scenario that fixed the trainer's consistency level to the true value.

We aim to test various settings as well as evaluating the algorithm in real-world environments. For the further extensions of the algorithm,
the limitation of the proposed approach is that the state and action spaces must be discrete. Although we could discretise the continuous state-action space, it might lead to a lager number of states and actions, limiting the available feedback for each action and state pair. 
Another interesting extension would be to estimate the consistency level for groups of states separately, rather than averaged over all states. It should identify variation of a trainer's consistency level over states. For example, this could be useful to detect a trainer who becomes unreliable only in certain situations. 

\small
\begin{ack}
This work is supported by SPHERE Next Steps Project funded by the UK's Engineering, and Physical Sciences Research Council (EPSRC) under Grant EP/R005273/1 and the UKRI Turing AI Fellowship EP/V024817/1.
\end{ack}

{
\small
\bibliographystyle{abbrv.bst}
\bibliography{my_library}

\begin{thebibliography}{10}

\bibitem{Bailer-Jones2011}
C.~Bailer-Jones and K.~Smith.
\newblock Combining probabilities.
\newblock {\em Data Processing and Analysis Consortium (DPAS),
  GAIA-C8-TN-MPIA-CBJ-053}, 2011.

\bibitem{Cederborg2015}
T.~Cederborg, I.~Grover, C.~L. Isbell, and A.~L. Thomaz.
\newblock Policy shaping with human teachers.
\newblock In {\em IJCAI}, pages 3366--3372, 2015.

\bibitem{Dempster1977}
A.~P. Dempster, N.~M. Laird, and D.~B. Rubin.
\newblock Maximum likelihood from incomplete data via the em algorithm.
\newblock {\em Journal of the royal statistical society. Series B
  (methodological)}, pages 1--38, 1977.

\bibitem{Griffith2013}
S.~Griffith, K.~Subramanian, J.~Scholz, C.~L. Isbell, and A.~L. Thomaz.
\newblock Policy shaping: Integrating human feedback with reinforcement
  learning.
\newblock In {\em Advances in neural information processing systems}, pages
  2625--2633, 2013.

\bibitem{Knox2012}
W.~B. Knox and P.~Stone.
\newblock Reinforcement learning from simultaneous human and mdp reward.
\newblock In {\em Proceedings of the 11th International Conference on
  Autonomous Agents and Multiagent Systems-Volume 1}, pages 475--482.
  International Foundation for Autonomous Agents and Multiagent Systems, 2012.

\bibitem{ni}
Y.~Ni, M.~McVicar, R.~Santos-Rodríguez, and T.~De~Bie.
\newblock Understanding effects of subjectivity in measuring chord estimation
  accuracy.
\newblock {\em IEEE Transactions on Audio, Speech, and Language Processing},
  21(12):2607--2615, 2013.

\bibitem{Pilarski2011}
P.~M. Pilarski, M.~R. Dawson, T.~Degris, F.~Fahimi, J.~P. Carey, and R.~S.
  Sutton.
\newblock Online human training of a myoelectric prosthesis controller via
  actor-critic reinforcement learning.
\newblock In {\em Rehabilitation Robotics (ICORR), 2011 IEEE International
  Conference on}, pages 1--7. IEEE, 2011.

\bibitem{raykar}
V.~C. Raykar, S.~Yu, L.~H. Zhao, G.~H. Valadez, C.~Florin, L.~Bogoni, and
  L.~Moy.
\newblock Learning from crowds.
\newblock {\em Journal of Machine Learning Research}, 11(43):1297--1322, 2010.

\bibitem{Tenorio2010}
A.~C. Tenorio-Gonzalez, E.~F. Morales, and L.~Villaseñor-Pineda.
\newblock Dynamic reward shaping: training a robot by voice.
\newblock In {\em Ibero-American Conference on Artificial Intelligence}, pages
  483--492. Springer, 2010.

\bibitem{Thomaz2008}
A.~L. Thomaz and C.~Breazeal.
\newblock Teachable robots: Understanding human teaching behavior to build more
  effective robot learners.
\newblock {\em Artificial Intelligence}, 172(6-7):716--737, 2008.

\bibitem{Watkins1989}
C.~J. C.~H. Watkins.
\newblock {\em Learning from delayed rewards}.
\newblock Phd thesis, 1989.

\end{thebibliography}
}

\clearpage
\appendix

\section{Appendix}

\subsection{M-step derivation}
\label{app:m-step}
The $i$-th iteration of the M-step for the EM algorithm can be derived by computing the following $\arg\max$ operation.
\begin{equation} \label{eq:mstep_1}
C^{\left(i+1\right)}_{s,a} = \arg\max_{C_{s,a}} \sum_{\mathcal{O}_{s,a}}P\left( \mathcal{O}_{s,a} | h_{s,a}^{+}, h_{s,a}^{-};C^{\left(i\right)}_{s,a})\right) \cdot \log\left( P\left( h_{s,a}^{+}, h_{s,a}^{-} | \mathcal{O}_{s,a};C_{s,a}\right)\right)
\end{equation}
Where $C^{\left(i+1\right)}_{s,a}$ is an estimated consistency level at $i$-th iteration. We use the following binomial distributions:
\begin{equation} \label{eq:mstep_2}
P\left( h_{s,a}^{+}, h_{s,a}^{-} | \mathcal{O}_{s,a} ; C_{s,a})\right) = 
	\begin{cases}
		\begin{pmatrix} h_{s,a}^{+} + h_{s,a}^{-} \\ h_{s,a}^{+} \end{pmatrix} C_{s,a}^{h_{s,a}^{+}} \left(1-C_{s,a}\right)^{h_{s,a}^{-}}
				& \mbox{if } \mathcal{O}_{s,a} = 1 \\
		\begin{pmatrix} h_{s,a}^{+} + h_{s,a}^{-} \\ h_{s,a}^{-} \end{pmatrix} \left(1-C_{s,a}\right)^{h_{s,a}^{+}} C_{s,a}^{h_{s,a}^{-}}
				& \mbox{if } \mathcal{O}_{s,a} = 0.
	\end{cases}
\end{equation}
By substituting Eq.\ref{eq:mstep_2} into Eq.\ref{eq:mstep_1} and computing the $\arg\max$, we obtain the following M-step:
\begin{equation} \label{eq:mstep_3}
C_{s,a}^{\left(i+1\right)} = \frac{P_{1} \cdot h_{s,a}^{+} + P_{0} \cdot h_{s,a}^{-}}{h_{s,a}^{+} + h_{s,a}^{-}},
\end{equation}
where $P_0$ and $P_1$ are given as follows:
\begin{equation} \label{eq:mstep_4}
\begin{split}
P_0 = P\left( \mathcal{O}_{s,a}=0 | h_{s,a}^{+}, h_{s,a}^{-};C_{s,a}^{\left(i\right)}\right), \\
P_1 = P\left( \mathcal{O}_{s,a}=1 | h_{s,a}^{+}, h_{s,a}^{-};C_{s,a}^{\left(i\right)}\right).
\end{split}
\end{equation}

\subsection{Consistency level estimation algorithm summary}
\label{app:const_est_summary}
\begin{algorithm}
\caption{Consistency Level Estimation}
\label{alg1}
\algsetup{
linenosize=\small,
linenodelimiter=.
}
\begin{algorithmic}[1]
\REQUIRE $i_{max}$, $P_{1}^{Q}(s,a)$, $h_{s,a}^{+}$ and  $h_{s,a}^{-}$ 
\STATE $\Delta(s,a) \leftarrow h_{s,a}^{+} - h_{s,a}^{-}$
\STATE $i \leftarrow 1$
\STATE $C^{(i)} \leftarrow 0.5$
\WHILE{TRUE}
  \STATE $P_0 \leftarrow \sum_{a'\neq a} P_{1}^{Q}(s,a') \cdot \left(C_{s,a'}^{(i)}\right)^{\Delta(s,a')}\left(1-C_{s,a'}^{(i)}\right)^{\sum_{j\neq a'}\Delta(s,j)} / Z$
  \STATE $P_1 \leftarrow P_{1}^{Q}(s,a) \cdot \left(C_{s,a}^{(i)}\right)^{\Delta(s,a)} \left(1-C_{s,a}^{(i)}\right)^{\sum_{j\neq a}\Delta(s,j)} / Z$
  \STATE $C^{\left(i+1\right)} \leftarrow \frac{P_{1} \cdot h_{s,a}^{+} + P_{0} \cdot h_{s,a}^{-}}{h_{s,a}^{+} + h_{s,a}^{-}}$
  \IF{$C^{(i+1)} == C^{(i)}$ or $i == i_{max}$}
    \STATE break
 \ENDIF
 \STATE $i \leftarrow i+1$
\ENDWHILE
\RETURN $C^{(i)}$
\end{algorithmic}
\end{algorithm}

\subsection{Adaptive learning rate derivation}
The generalised variance of the updated consistency level $\sigma'^2$ is defined as,
\begin{equation} \label{eq:prec_def1_a}
\sigma'^2 = 1/\lambda' = \mathbb{E}\left[\left(C' - C^*\right)^2\right].
\end{equation}
The consistency level update equation is,
\begin{equation} \label{eq310a}
C' = C + \alpha \cdot ( C(s,a) - C ).
\end{equation}
By substituting Eq.~\ref{eq310a} to Eq.~\ref{eq:prec_def1_a}, we obtain,
\begin{equation} \label{eq:prec_def2_a}
\begin{split}
\sigma'^2 &= \mathbb{E}\left[\left( \left(1-\alpha\right)\left(C-C^*\right) + \alpha\left(C(s,a)-C^*\right)\right)^2\right] \\
&= \left(1-\alpha\right)^2\mathbb{E}\left[(C-C^*)^2\right] + \alpha^2\mathbb{E}\left[(C(s,a)-C^*)^2\right] \\
&= \left(1-\alpha\right)^2\sigma^2 + \alpha^2\sigma_{s,a}^2.
\end{split}
\end{equation}
In above, we assumed the error in $C$ and $C(s,a)$ are statistically independent.
We derive the $\alpha$ that minimises $\sigma'^2$ by solving $\frac{\partial\sigma'^2}{\partial\alpha}=0$,
\begin{equation} \label{eq:adap_lr_alpha_a}
\alpha = \frac{\sigma^2}{\sigma_{s,a}^2+\sigma^2} = \frac{\lambda_{s,a}}{\lambda_{s,a}+\lambda}.
\end{equation}
Then by substituting Eq.~\ref{eq:adap_lr_alpha_a} back to Eq.~\ref{eq:prec_def2_a}, we get following updated precision,
\begin{equation} \label{eq:adapt_lr_lambda_a}
\lambda' = \lambda_{s,a}+\lambda,
\end{equation}

\subsection{Adaptive learning rate algorithm summary}
\label{app:adaptive_learning_rate_summary}
\begin{algorithm}
\caption{Adaptive Learning Rate for Consistency Level Estimation}
\label{alg2}
\begin{algorithmic}[1]
\REQUIRE $\zeta$, $C(s,a)$ and $\lambda_{s,a}$
\REQUIRE $C$, $\lambda$ persistent variables ($C$ initialized 0.5, $\lambda$ initialized 0 or small positive number)
\STATE $\alpha \leftarrow \frac{ \lambda_{s,a} }{ \lambda_{s,a} + \lambda }$
\STATE $C \leftarrow C + \alpha \cdot ( C(s,a) - C )$
\STATE $\lambda \leftarrow \zeta\lambda + \lambda_{s,a}$
\RETURN $C$
\end{algorithmic}
\end{algorithm}

\subsection{Experiment details}
\label{app:exp_details}
\subsubsection{Environment}
\begin{figure} [H]
\begin{center}
\includegraphics[width=6cm]{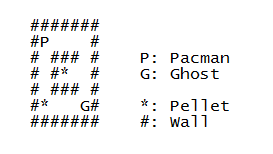}
\caption{Screenshot of 5x5 grid world Pac-Man}
\label{fig:pacman}
\end{center}
\end{figure}

Fig.~\ref{fig:pacman} shows the starting state of the game.

\subsubsection{Constructing an oracle}
Similar to \cite{Griffith2013}, we also use Oracle to simulate human trainer feedback because it allows us to sweep parameters of feedback likelihood ($L$), consistency level ($C$) and the number of trainers. The Oracle was created by using Q-learning in the environment. After 10,000 episodes, we confirmed that the averaged reward converged, and it always clears the game by checking the total reward value, then store the Q-function table. The oracle uses the stored Q-function table with a greedy policy to generate the optimum action on a given state. Then the feedback is generated based on this optimum action in conjunction with feedback likelihood ($L$) and consistency level ($C$.)  

\subsubsection{Underlying Reinforcement Learning algorithm}
We need an \ac{RL} algorithm to learn an environment from interaction with it, as well as algorithm learning from human feedback. In this experiment, we use the Q-Learning algorithm with Boltzman exploration policy for the underlying \ac{RL} algorithm. 
Q-Learning algorithm uses the following one step time difference update for estimating Q-function.
\begin{equation}
Q\left(s_{t},a_{t}\right) = Q\left(s_{t},a_{t}\right) + \alpha_{Q} \left( r_{t} + \gamma \max_{a'} Q \left(s_{t+1}, a'\right) - Q\left(s_{t},a_{t}\right) \right) 
\end{equation}
Where $\alpha_{Q}$ is learning rate, and $\gamma$ is discount factor. For all of our experiments, we used $\alpha_{Q} = 0.05$ and $\gamma = 0.9$.

Boltzman exploration policy uses the following probability to take action 'a' in state 's'.

\begin{equation}
\pi_{R}\left(s,a\right) = \frac{ e^{Q\left(s,a\right)} / \tau }
                                          { \sum_{a' \in \mathcal{A}} e^{Q\left(s,a'\right)} / \tau }
\end{equation}
Where $\mathcal{A}$ is a group of all possible actions and $\tau$ is the temperature parameter that controls the balance between exploration and exploitation, and it is fixed to 1.5 for all of our experiments.

For estimating the consistency level, the underlying \ac{RL} algorithm must generate the probabilities of given state-action pair (s,a) is optimum $P_1^{Q}$ and not optimum $P_0^{Q}(s,a)$. We use the following for them.
\begin{equation}
\begin{split}
P_1^{Q}(s,a) &= \pi_{R}(s,a) \\
P_0^{Q}(s,a) &= 1 - \pi_{R}(s,a)
\end{split}
\end{equation}

\subsubsection{Results}
All total rewards are averaged over 200 learning trials. Each trial has 2000 episodes and is then applied to smooth (moving average) over 21 episodes window length. All the consistency level estimations are averaged over 200 learning trials, each trial has 2000 episodes, and no smoothing is applied.

\acrodef{RL}{reinforcement learning}
\acrodef{DRL}{deep reinforcement learning}
\acrodef{DQN}{deep Q-Learning}
\acrodef{DNN}{deep neural network}
\acrodef{MDP}{Markov decision process}

\end{document}